\documentclass[conference]{IEEEtran}
\IEEEoverridecommandlockouts
\usepackage{cite}
\usepackage{amsmath,amssymb,amsfonts}
\usepackage{algorithmic}
\usepackage{graphicx}
\usepackage{svg}
\usepackage{textcomp}
\usepackage{xcolor}

\usepackage[
  letterpaper,
  top=1in,
  bottom=1in,
  left=0.75in,
  right=0.75in
]{geometry}

\newcommand{\mycomment}[1]{}
\def\BibTeX{{\rm B\kern-.05em{\sc i\kern-.025em b}\kern-.08em
    T\kern-.1667em\lower.7ex\hbox{E}\kern-.125emX}}

\begin{document}
\title{A Causal Probabilistic Framework for Perception-Informed Closed-Loop Simulation of Autonomous Driving}
\author{
    \IEEEauthorblockN{
        Zhennan Fei\textsuperscript{1,4}, 
        Rickard Johansson\textsuperscript{1}, 
        Mikael Andersson\textsuperscript{1},
        Matthias Eng\textsuperscript{1}, 
        Mattias Eriksson\textsuperscript{1}, \\
        Kaveh Kianfar\textsuperscript{1},
        Sadegh Rahrovani\textsuperscript{1}, 
        Chris van der Ploeg\textsuperscript{2,5},
        Michael Borth\textsuperscript{2},
        Maren Buermann\textsuperscript{2}, \\
        Michiel Braat\textsuperscript{2},
        Henk Goossens\textsuperscript{2},
        Zijian Han\textsuperscript{3},
        Majid Khorsand Vakilzadeh\textsuperscript{3},
        Gabriel Rodrigues de Campos\textsuperscript{3}
    }
    \IEEEauthorblockA{\textsuperscript{1}Volvo Cars, Sweden}
    \IEEEauthorblockA{\textsuperscript{2}TNO, The Netherlands}
    \IEEEauthorblockA{\textsuperscript{3}Zenseact, Sweden}
    \IEEEauthorblockA{\textsuperscript{4}Chalmers University of Technology, Sweden}
    \IEEEauthorblockA{\textsuperscript{5}Eindhoven University of Technology, The Netherlands}
}
\maketitle

\begin{abstract}
Software-in-the-loop (SIL) simulation is a cornerstone for the validation of modern automotive safety functions. However, many current frameworks utilize "ideal sensing," which bypasses the functional insufficiencies of perception algorithms, leading to over-optimistic safety assessments. This paper proposes a perception-informed SIL testing methodology that bridges the gap between ground-truth simulation and real-world perception behavior. We present a framework for incorporating causal probabilistic models into standardized, scenario-based simulation toolchains, applicable to both Advanced Driver Assistance Systems (ADAS) and Autonomous Driving Systems (ADS). Our approach enables the systematic injection of realistic perception errors—such as loss of detection, sizing inaccuracies, and positioning offsets—derived from physical triggering conditions like fog, rain, and object-merging scenarios. By evaluating these "faults" within a standardized simulation environment, we demonstrate that perception-informed testing reveals latent operational risks that ideal SIL environments fail to capture, providing a scalable pathway for SOTIF (ISO 21448) validation.
\end{abstract}


\section{Introduction}
The advancement of Advanced Driver Assistance Systems (ADAS) and Autonomous Driving (AD) functions represents a transformative shift in the automotive industry, aiming to significantly reduce traffic accidents and improve transportation efficiency. As these systems assume increasing responsibility for the dynamic driving task, ensuring their safety and reliability under diverse operational conditions becomes a paramount challenge. Traditionally, validation and verification (V\&V) have focused on functional safety as defined by ISO 26262~\cite{iso26262}, which addresses risks from E/E system malfunctions. However, modern safety assurance must also encompass the Safety of the Intended Functionality (SOTIF) under ISO 21448~\cite{iso21448}. SOTIF focuses on mitigating hazardous events caused by functional insufficiencies—limitations in the performance of perception or decision-making algorithms—that occur even in the absence of a system failure.

Software-in-the-loop (SIL) simulation has become the cornerstone of ADAS and AD development due to its scalability and reproducibility. However, a critical fidelity gap exists in most current SIL frameworks that utilize object-level sensor models. In these environments, the simulation's ground truth is typically filtered by a basic sensor view—applying geometric constraints such as field-of-view (FOV) and occlusion—to generate an object list. While this approach models the visibility of actors, the resulting data remains essentially "ideal" because it lacks the physical and environmental noise that characterizes real-world sensing. Specifically, this method fails to account for the causal chain between external triggering conditions—such as degraded contrast in heavy fog or digitalization noise in low-light environments—and the resulting functional insufficiencies in the perception stack. Consequently, such ideal sensing setups cannot validate system robustness against perception degradation or temporary failures, a requirement of increasing focus for safety rating organizations like EuroNCAP~\cite{euroncap}, which demands evidence of reliability under obscured or non-ideal conditions.

To bridge this fidelity gap, ``full-stack'' SIL simulation can be employed, where raw sensor data (e.g., camera pixels) is rendered and processed by the actual perception software. While full-stack simulation is an invaluable tool for debugging rare corner cases during development, it is computationally expensive and requires significant GPU resources. 
This intensity makes full-stack SIL unsuitable for the massive-scale simulation and large permutations of parameters demanded by emerging regulatory and rating frameworks, such as UN Regulation No. 157 (ALKS)~\cite{alks}, the Driver Control Assistance Systems (DCAS)~\cite{dcas} regulations, or the comprehensive scenario catalogs of EuroNCAP~\cite{euroncap}. There is a critical need for a methodology that captures realistic perception failures with the computational efficiency of object-level simulation.

In this paper, we propose a perception-informed SIL testing framework that utilizes causal probabilistic modeling to bridge this gap. We focus specifically on the camera sensor, which is highly susceptible to environmental triggers and serves as a primary modality for most current ADAS and ADS. By employing Bayesian Networks~\cite{pearl1988}, we characterize camera-specific failure modes—including loss of detection (missed detections), object sizing inaccuracies, and positioning errors (lateral and longitudinal)—and map them to causal factors in the Operational Design Domain (ODD). Unlike open-loop perception analysis, our framework operates in a closed-loop manner: the probabilistic perception errors are injected back into the software pipeline, directly influencing the vehicle's control logic and subsequent dynamics.

The proposed framework is realized within a specialized, in-house simulation toolchain designed for scalable validation of ADAS/ADS. This toolchain integrates esmini~\cite{esmini}, an open-source OpenSCENARIO interpreter, for scenario execution and world modeling with an in-house vehicle dynamics model. Within this architecture, the probabilistic model operates as a statistical inference engine: it takes the object-level outputs from an ideal camera sensor model as inputs and performs real-time statistical reasoning based on the current ODD parameters. By systematically injecting faults as perception insufficiencies directly into the object stream, the framework enables the observation of their impact on closed-loop ego-vehicle behavior. This provides a portable, high-throughput toolchain that achieves realistic safety assessments without the prohibitive GPU overhead and computational costs of raw data rendering.

The contributions of this work are as follows:

\begin{itemize}
    \item We develop a Bayesian probabilistic framework mapping environmental triggering conditions to three specific camera perception functional insufficiencies.
    \item We present an integrated, closed-loop SIL toolchain based on open standards that enables high-scale robustness testing by feeding probabilistic perception faults back into the vehicle control system.
    \item We demonstrate through safety-critical scenarios that this perception-informed framework enables the rigorous testing of decision and control logic robustness, identifying hazardous reactions to functional insufficiencies that remain hidden in traditional "ideal" SIL environments.
\end{itemize}

\section{Related Work}
The validation of ADAS/AD functions relies heavily on simulation to bridge the gap between development and real-world deployment. Related research in this domain can be broadly categorized into full-stack sensor simulation and probabilistic perception error modeling.

Many modern testing frameworks, such as~\cite{FullStackSimFailureSim}\cite{EnvRecognitionCarla}\cite{SimsV}, utilize high-fidelity, closed-loop simulation to validate system behavior by rendering raw sensor data. Specifically, Matos et al.~\cite{FullStackSimFailureSim} presents a simulation-based fault injection framework using the CARLA~\cite{Carla} simulator integrated with an open-source autonomous driving stack to evaluate the impact of sensor failures—specifically LiDAR, GNSS, and IMU on system-level safety. The study demonstrates that complete failures or severe noise in LiDAR and IMU gyroscope data have the most critical impact, often leading to erratic vehicle motion and collisions in closed-loop urban driving scenarios. Gorza et al.~\cite{EnvRecognitionCarla} investigates the performance of an environment recognition system for autonomous vehicles using CARLA and the YOLOv7 object detection model. By simulating various urban traffic scenarios under challenging weather and lighting conditions (e.g., night, heavy rain, and extreme fog), the authors evaluate detection reliability across multiple object classes, identifying significant performance degradations for smaller or less distinct objects like traffic signs and lights. Furthermore, Pham et al.~\cite{SimsV} proposes SimsV, a perception-guided fuzzing technique designed for system-level testing of multi-module ADS. SimsV leverages high-fidelity simulation to continuously apply mutation operators to driving scenarios, effectively finding perception weaknesses that lead to severe safety issues, including collisions, by exploiting the impact of perception failures on the system's overall decision-making. While these high-fidelity approaches effectively reveal perception weaknesses and their system-level consequences by rendering raw sensor data, but they require prohibitive GPU resources and precise sensor models that are computationally expensive to execute at scale. 

To address the computational bottlenecks of full-stack simulation, researchers have explored the use of stochastic models~\cite{DriveFI}\cite{Gansch26}\cite{PEM} to inject failures at the object list level. In~\cite{DriveFI}, a fault-injection framework referred to as DriveFI is proposed to utilizes Bayesian Networks (BN) to mine internal accidental faults—such as GPU bit-flips—that impact vehicle safety in end-to-end closed-loop simulations. Similar to our framework, DriveFI utilizes a closed-loop simulation to quantify safety via a ``safety potential'' metric based on stopping distances and safety envelopes. However, our work differs fundamentally in its failure source and technical objective. While DriveFI focuses on accidental faults (Functional Safety, ISO 26262) arising from internal system malfunctions, our framework addresses functional insufficiencies (SOTIF, ISO 21448) triggered by external environmental conditions such as fog and rain. 


Gansch et al.~\cite{Gansch26} share the theoretical foundation with our work, utilizing Bayesian Networks to move beyond purely associative safety reasoning. Their approach aligns with SOTIF as it explicitly models environmental triggering conditions—such as occlusion, object size, and traffic density—to quantify risk propagation within perception subsystems. Diverging in objective and scope, Gansch et al. focus on safety analysis and importance ranking metrics for perception components, whereas our work focuses on the robustness testing of the full closed-loop control system.

The Perception Error Model (PEM) framework proposed in~\cite{PEM} exhibits several similarities to our approach, notably the use of a virtual component to inject stochastic perception errors into object-level data within a closed-loop simulation. Both frameworks share the common objective of evaluating how these injected ``faults'' propagate through the system to impact the ego-vehicle's downstream decision-making and control logic. However, a fundamental difference lies in the underlying modeling philosophy: while PEM adopts a data-driven "black-box" approach that derives error distributions by statistically comparing ground-truth data to perception outputs, our framework utilizes a causal, physics-informed approach. Specifically, we explicitly map environmental triggering conditions to perception failures using Bayesian Networks, whereas PEM relies on statistical object attributes like distance and orientation. This causal methodology is specifically tailored for SOTIF (ISO 21448) validation, as it establishes the direct causal chain between Operational Design Domain (ODD) factors and functional insufficiencies.

\section{Proposed Method}
In this section, we present the development of the proposed framework designed to quantify the functional insufficiencies of camera-based perception. The methodology first establishes the mathematical foundations of Bayesian Networks (BNs). We then describe the construction of a structural causal model where environmental triggering conditions are mapped to perception performance metrics like contrast and sharpness, etc,. Following the systematic approach proposed in~\cite{borth_2024,borth_2025}, which is in turn based on~\cite{Borth_2002}, the BN structure is derived from technical system specifications and Failure Mode and Effects Analysis (FMEA). Specifically, we model the information flow of the camera-based perception where, as the information propagates through the environment, it is degraded by physical effects (e.g., rain and fog) as well as system specific effects (loss of light in the lens, digitalization effects). This ensures that the model captures known functional dependencies and failure modes. As a proof-of-concept, the model utilizes physical relationships and engineering assumptions to reason how environmental states lead to functional insufficiencies without requiring immediate dataset calibration. Finally, we detail the integration of this probabilistic engine into the in-house simulation architecture.

\subsection{Bayesian Networks for Continuous Variables}
A Bayesian Network (BN) is a probabilistic graphical model that represents a set of random variables and their conditional dependencies via a Directed Acyclic Graph (DAG). In the context of autonomous driving perception, the variables of interest—such as illumination levels, object distances, and image contrast—are inherently continuous.

Mathematically, a BN is defined by a pair $(G, \theta)$, where $G = (V, E)$ is a Directed Acyclic Graph (DAG) with nodes $V$ representing random variables $\{X_{1}, X_{2}, \dots, X_{n}\}$ and edges $E$ representing direct causal influences. The parameter set $\theta$ defines the Conditional Probability Distributions (CPDs) for each node. For continuous variables, the joint probability density function $f(x_{1}, \dots, x_{n})$ is factorized according to the chain rule of Bayesian networks:

\begin{equation}
f(x_{1}, \dots, x_{n}) = \prod_{i=1}^{n} f(x_{i} \mid pa(X_{i}))
\end{equation}

where $pa(X_{i})$ denotes the set of parent nodes of $X_{i}$ in the graph $G$.

The relationship between a child node and its parents is typically modeled using Conditional Gaussian distributions or functional relationships with additive noise:

\begin{equation}
X_{i} = g_{i}(pa(X_{i})) + \epsilon_{i}
\end{equation}

where $g_{i}$ represents a physical or empirical function (e.g., the relationship between light intensity and image contrast) and $\epsilon_{i}$ represents Gaussian noise accounting for model uncertainty or sensor stochasticity. 

In this study, the Bayesian network is implemented using aGrUM~\cite{agrum_2017}. Consequently, the prior distributions and CPDs are converted into their discrete versions, making the selection of state-space size and resolution a key determinant of the model’s precision and accuracy, as well as of the computational cost of the Bayesian network. 

\subsection{Causal Model Construction}
We build the causal model that maps real-world external (e.g., from the environment) and internal triggering conditions to specific perception performance metrics. This model allows us to track how physical factors degrade the image as it moves through the ``optical pathway''. Here, the main attributing physical principles are related to the loss of contrast and a lack of spatial frequency (i.e., sharpness).
By formalising these steps, we can predict how these physical stressors might lead to a loss of information and ultimately could result in a failure or functional insufficiency of the perception system.

\subsubsection{Contrast reasoning: from contrast loss to misdetection}

\begin{figure}[t]
    \centering
    \includegraphics[width=11.2cm]{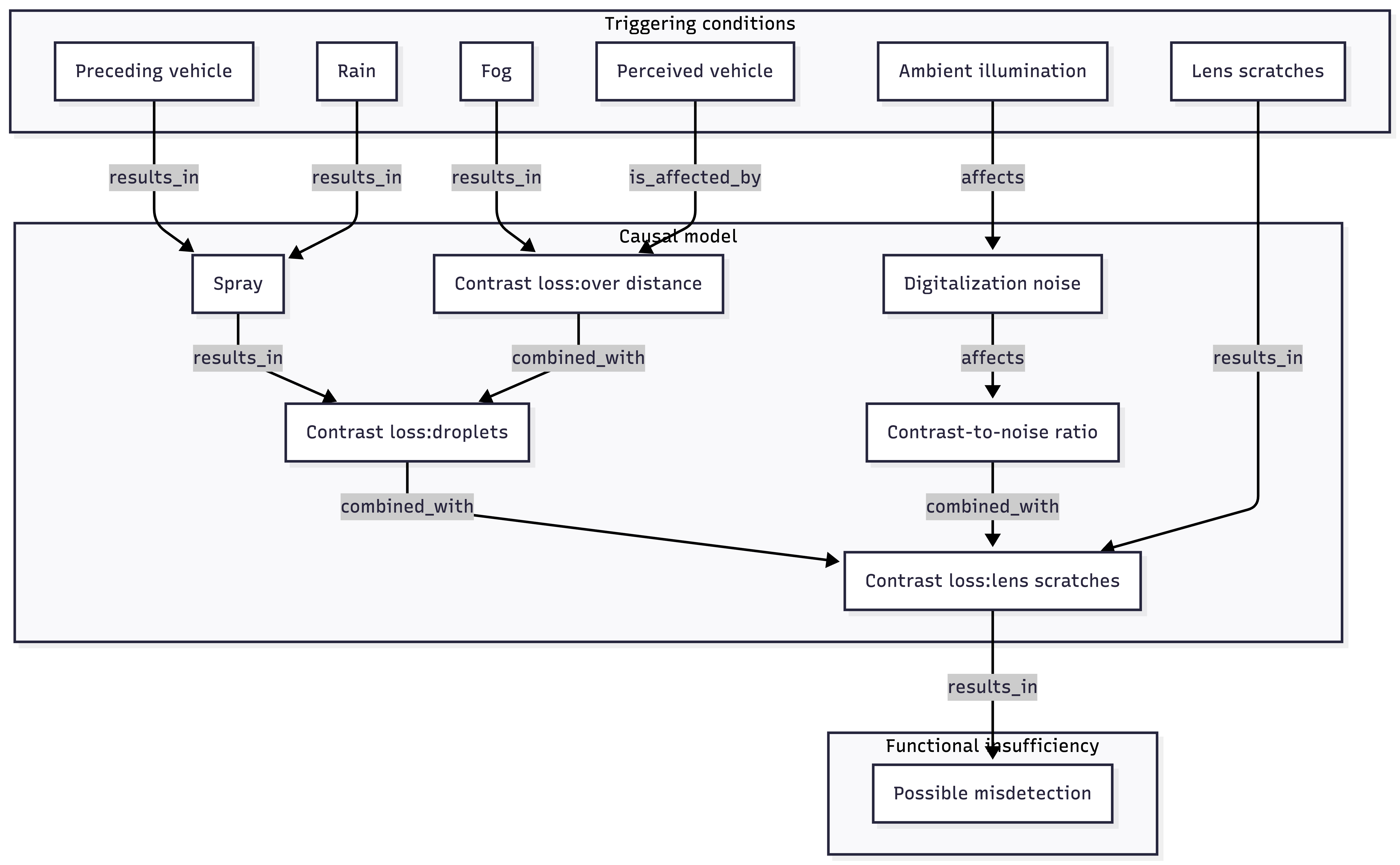}
    \caption{Causal chain of contrast degradation and detection failure}
    \label{fig:causal_model_contrast}
\end{figure}

To evaluate how well the perception system performs, we model the ``loss'' of contrast as light travels from an object to the camera sensor. It starts with an initial contrast of the object w.r.t. its environment (or lack thereof, e.g., a black car in front of a black wall). As the contrast decreases, the probability that the system will successfully detect the object also drops. Figure~\ref{fig:causal_model_contrast} illustrates the causal relationships used in our Bayesian Network to estimate this degradation.

\begin{figure}[b]
    \centering
    \includegraphics[width=0.9\columnwidth]{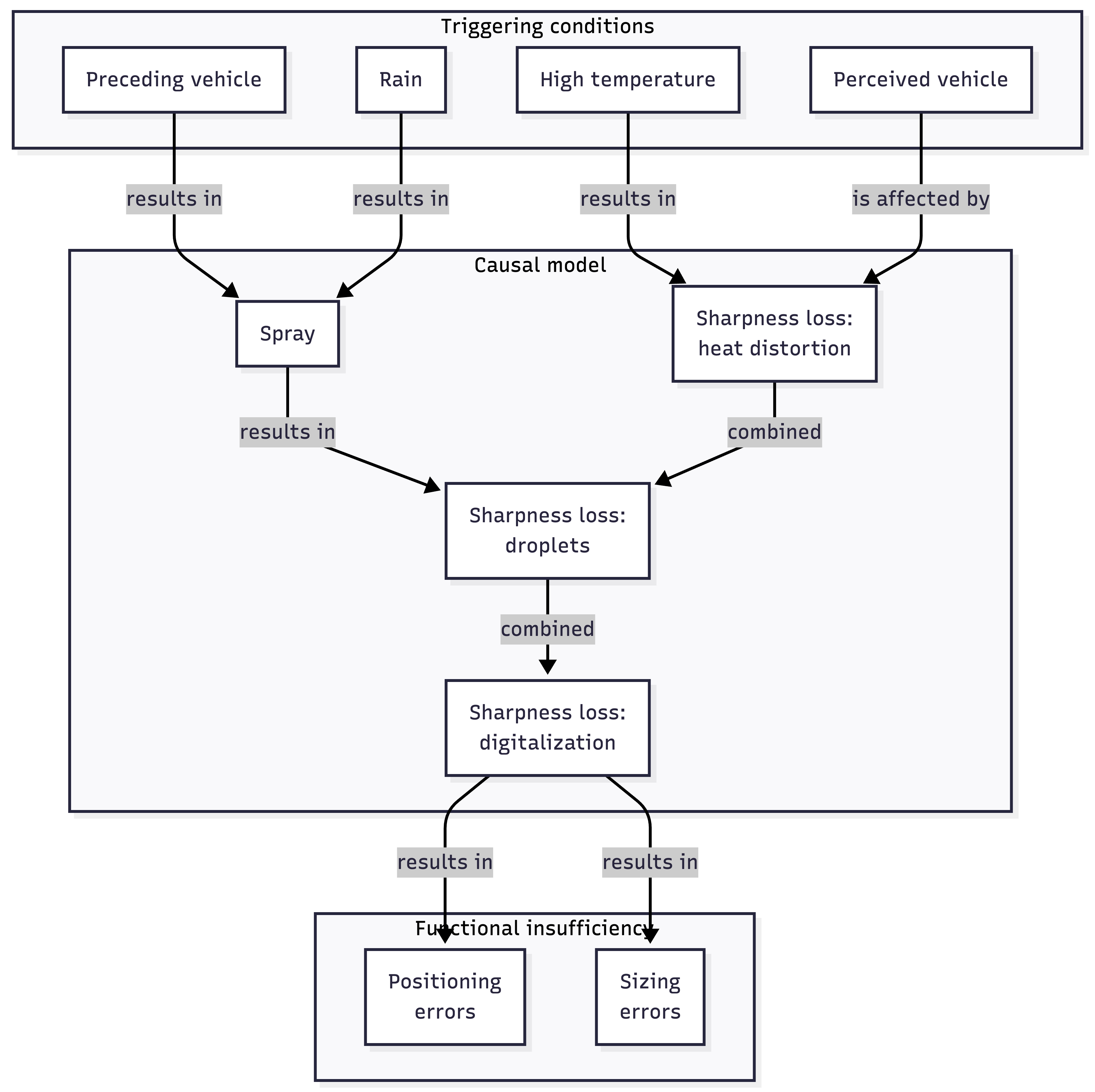}
    \caption{Causal chain of sharpness degradation and localization errors}
    \label{fig:causal_model_sharpness}
\end{figure}

\begin{figure}[t]
    \centering
    \includegraphics[width=0.5\columnwidth]{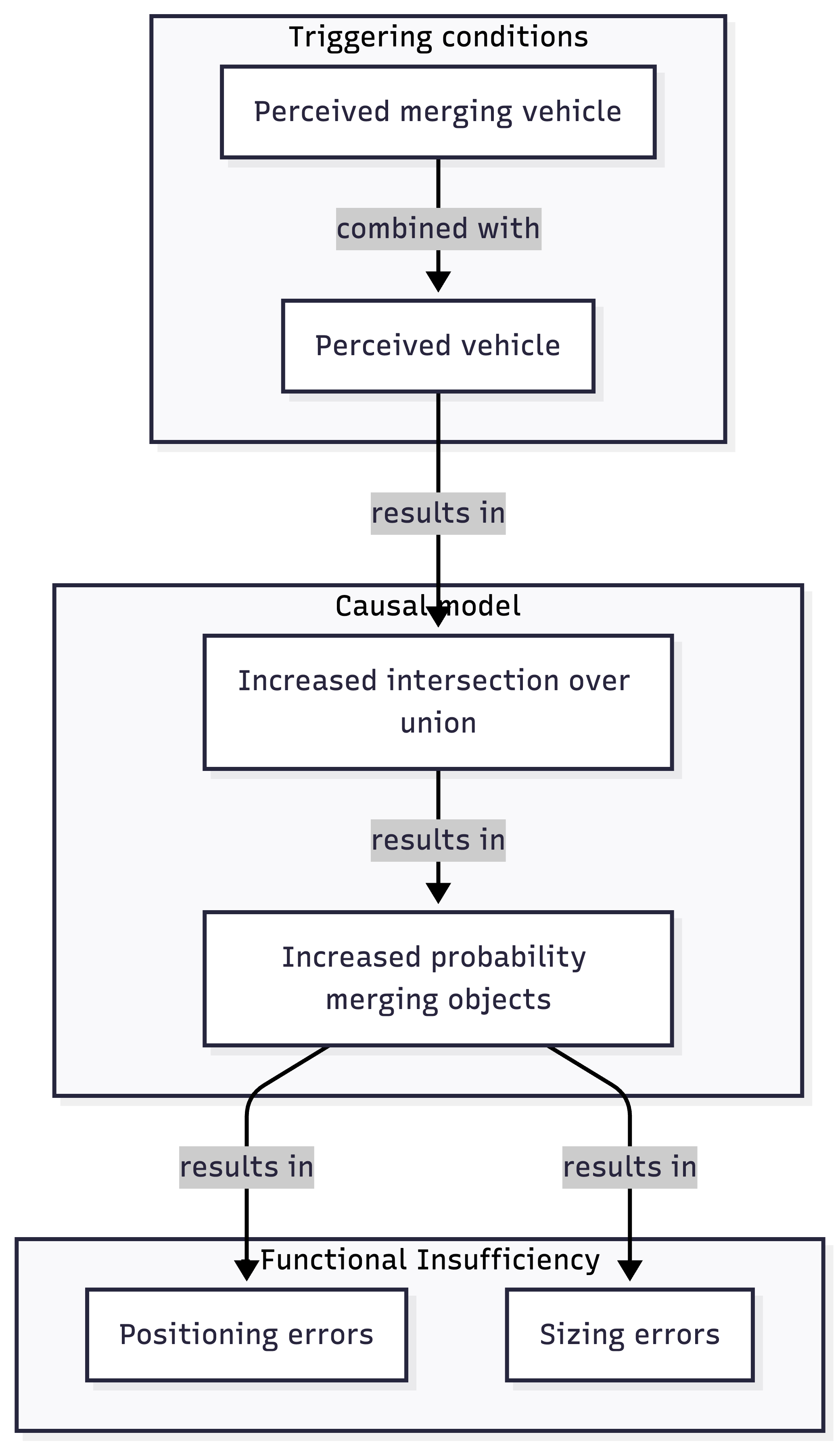}
    \caption{Causal chain of object merging and localization failures}
    \label{fig:causal_model_obj_merge}
\end{figure}

The model breaks down contrast degradation into three primary stages:
\begin{itemize}
    \item Physical contrast decay. The model traces information loss starting from the inherent contrast between the object and its background. As distance increases, fog introduces atmospheric scattering that causes contrast to decay. Additional loss occurs when rain or spray from preceding vehicles creates droplets on the lens, further scattering light. Finally, internal factors like lens scratches contribute to the cumulative degradation.
    \item Digitalization and noise. Once light reaches the sensor, ambient illumination and digitalization effects introduce noise. The model reasons primarily about the Contrast-to-Noise Ratio (CNR), as a low CNR negatively impacts the edge detection capabilities of neural networks, leading to degraded performance~\cite{edgedetection_2022}.
    \item To decide if a detection will fail, we compare the contrast and noise values to specific expectations of algorithm. Based on experimental validation and literature rules-of-thumb, we use the following benchmarks: (i) Normalized contrast: values below $10$ typically lead to detection failure, while values above $30$ are usually sufficient. (ii) CNR: a ratio below $3$ is considered insufficient for detection, whereas a ratio above $6$ is normally sufficient.
\end{itemize}

\subsubsection{Sharpness reasoning: from edge filtering to positioning and size errors}
A typical ADAS/AD functional requirement is to detect and locate the vehicle. Sharpness degradation directly violates this by causing positioning errors (lateral and longitudinal) and sizing errors. These errors are the result of distorted pixel-space information being fed into the underlying perception algorithms, such as neural networks.

In addition to contrast, we model the ``loss'' of sharpness as information traverses the optical pathway. Rather than a binary determination of whether an image is sharp enough, we characterize sharpness by a kernel—typically assumed to be Gaussian—that filters the image. This filtering process results in the perceived object appearing symmetrically or asymmetrically larger or smaller in pixel space, which ultimately translates to skewed object sizes and incorrect perceived distances.

Figure~\ref{fig:causal_model_sharpness} illustrates the causal progression from environmental stressors to these functional errors. The model identifies several stages of sharpness degradation that contribute to localization failures:
\begin{itemize}
    \item Physical filtering effects. Environmental factors create filtering effects that reduce spatial frequency. For instance, a high temperature difference between the ground and air creates a heat distortion filtering effect on the perceived object. Similarly, rain droplets on the lens create a diffusion effect that further filters the object's edges.
    \item Symmetric vs. asymmetric distortions. The nature of the stressor determines the type of distortion. Homogeneous rain or heat differences typically result in symmetric effects, while droplets partially obscuring an object can lead to asymmetric distortions.
    \item Digitalization and kernel spread. Beyond environmental factors, the camera sensor introduces a ``sharpness loss digitalization'' stage. The spread of the resulting sharpness kernel dictates the degree to which an object is perceived as larger or smaller.    
\end{itemize}

\subsubsection{Object merge reasoning: scenario-induced merging and localization Errors}

Object merging represents a perception failure mode, particularly in cut-in and cut-out scenarios where two physically distinct vehicles are incorrectly represented as a single detection. The causal logic for this failure is driven by the interaction between the scenario geometry and the design of the perception algorithm. In this context, scenario geometry refers to the spatial configuration and mutual physical positions of the vehicles—specifically their lateral and longitudinal offsets—which, when projected onto the 2D image plane, result in overlapping bounding boxes.

When objects are merged, the resulting single bounding box fails to accurately represent the spatial extent of either vehicle. Explicitly, this results in bounding box errors that implicitly also propagate to positioning errors and sizing errors, violating the functional requirement to accurately detect and locate individual vehicles in the scene.

Figure~\ref{fig:causal_model_obj_merge} outlines the causal progression from vehicle proximity to functional errors. The model identifies the following stages in the object merge causal chain.

\begin{itemize}
    \item Geometric triggering and IoU. The process begins when the mutual positions of two vehicles result in a high horizontal overlap of their bounding boxes in the image plane. This overlap is quantified by the Intersection-over-Union (IoU):
    \[
    \text{IoU} = \frac{|A \cap B|}{|A \cup B|}
    \]
    High IoU values create visual ambiguity, leading to an increased Intersection over Union node in the causal model.
    \item Algorithmic Filtering (NMS). To eliminate redundant detections, a typical perception system utilizes Non-Maximum Suppression (NMS). NMS sorts all detected boxes by their confidence scores and suppresses any box whose IoU with a higher-confidence box exceeds a threshold $T$. This threshold defines the limit at which overlapping cars are seen as two distinct detections. In our model, this also defines whether the objects are merged and, consequently, whether one of the two objects is removed from the ideal sensor measurement while the other is altered in size and position. As with the sharpness effect, any modification of the bounding box dimensions leads to a corresponding change in the perceived size and location of the object. The relative influence of these two effects depends on the specific system.
\end{itemize}

\subsection{Toolchain Integration and Probabilistic Simulation}

\begin{figure}[t]
    \centering
    \includegraphics[width=0.50\textwidth]{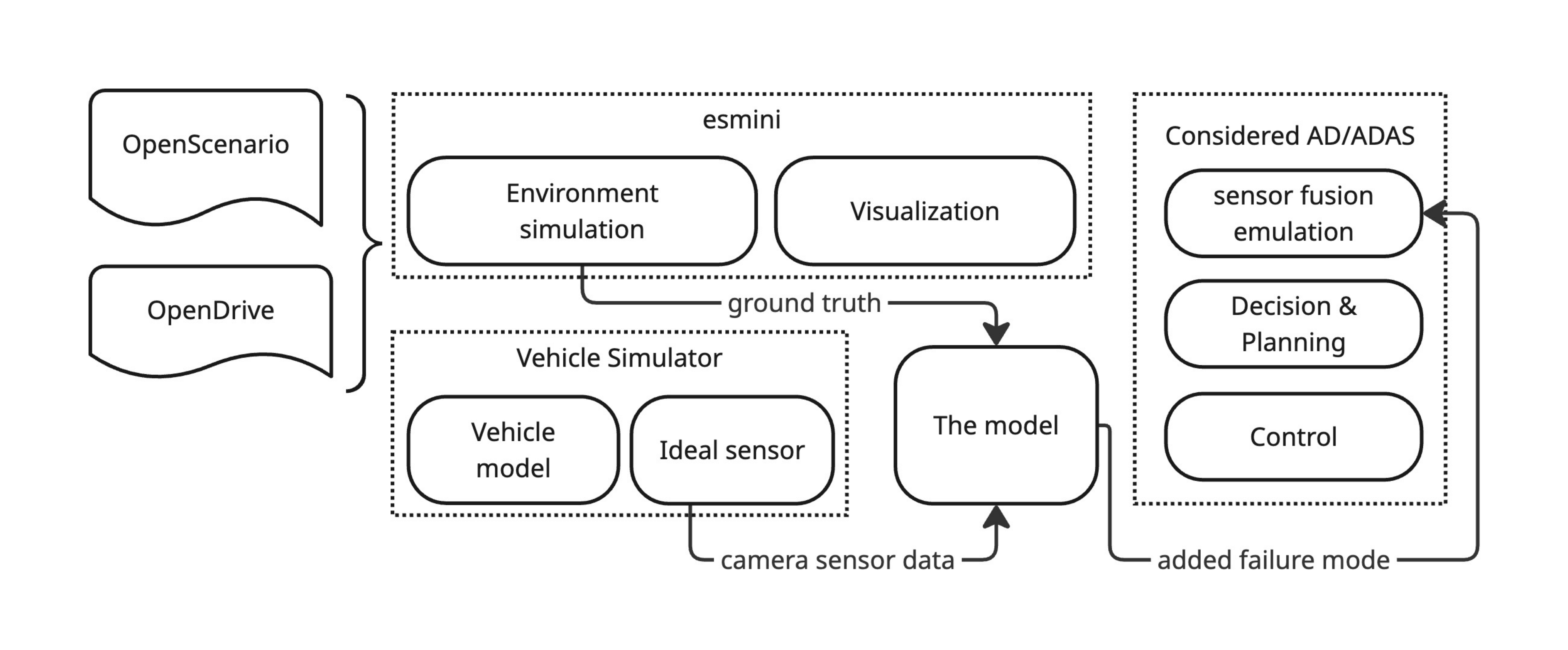}
    \caption{Integration architecture for failure injection in the SIL toolchain}
    \label{fig:sil_architecture}
\end{figure}

The proposed causal model is integrated into a Software-in-the-Loop (SIL) toolchain to evaluate the impact of perception-level insufficiencies on the considered ADAS/AD functions. The SIL simulation architecture, illustrated in Figure~\ref{fig:sil_architecture}, utilizes esmini for environment simulation and visualization, driven by standardized OpenScenario~\cite{asam-openscenario} and OpenDrive~\cite{asam-opendrive} inputs. A dedicated vehicle simulator provides a vehicle model and an ideal camera sensor, which generates data by transforming ground truth into the sensor's mounting position and applying geometric filters for field-of-view and occlusion.

The primary data interface for this toolchain is the Open Simulation Interface (OSI)~\cite{asam-osi}, which provides a hierarchical and standardized message format for environmental and object data. The model acts as a transformation layer, reading ground truth environmental conditions and object information from the simulation via OSI. During execution, the model performs Bayesian inference at each simulation time step, taking the observed OSI inputs as evidence to calculate the posterior probabilities of specific failure modes. Based on this statistical inference, the model stochastically modifies the object information within the OSI messages—injecting failure modes like positioning errors or merged detections—before passing the data to the sensor fusion emulation of the AD/ADAS stack. This allows the stack's decision-making and control modules to react to realistic perception failures rather than ideal data.

\section{Experimental Results}

    

The experimental evaluation discussed in this section serves a dual purpose: first, to validate the causal model and its integration within the SIL toolchain, and second, to demonstrate the utility of the model in injecting failure modes to evaluate the robustness of an ADAS function. 

\subsection{Experimental Scenarios}

The evaluation is conducted across three driving scenarios to demonstrate the practical application of the causal model within the SIL toolchain.

\emph{Car-to-Car Rear Stationary (CCRs) in Darkness.}
The first scenario 
focuses on a stationary preceding vehicle encountered under conditions of ambient illumination. This scenario is adopted to test the contrast reasoning path of the model, specifically evaluating how low ambient light levels affect digitalization noise and the resulting CNR. The objective is to observe possible misdetections as the CNR falls below established sufficiency thresholds. By injecting these failures into the SIL toolchain, we evaluate the robustness of ADAS functions when the ability to ``see'' a stationary object is compromised.

\emph{Car-to-Car Rear Moving (CCRm) in Fog and Rain.}
The second scenario 
involves a moving preceding vehicle under fog and rain. This configuration simultaneously activates multiple causal chains: (i) Fog intensity and distance induce a decay in contrast and a loss of sharpness due to atmospheric scattering. (ii) Rain and the resulting spray from the preceding vehicle create droplets on the lens, further degrading both contrast and sharpness.
The experiment evaluates how this optical pathway degradation translates into positioning and sizing errors, while simultaneously testing for misdetections triggered when the combined contrast loss from fog and spray drives the CNR below critical limits. By injecting these failures, we can assess the robustness of longitudinal control functions as they attempt to maintain safe operation despite a fluctuating or intermittently disappearing lead vehicle.

\emph{Cut-out.}
The final scenario 
involves a multi-vehicle interaction featuring the ego vehicle, a cut-out vehicle, and a vehicle ahead of it. This scenario is specifically designed to evaluate object merge reasoning. As the cut-out vehicle maneuvers, the model calculates the probability of a merge based on the horizontal overlap and the resulting increased Intersection over Union (IoU) between the two preceding objects. This scenario evaluates the ADAS stack's ability to navigate scenes where targets may suddenly merge, potentially leading to incorrect control responses.

The experimental scenarios and the associated environmental conditions are implemented using the Python package scenariogeneration~\cite{scenariogeneration}. The package automatically generates the OpenSCENARIO (.xosc) and OpenDRIVE (.xodr) XML files required as inputs for the SIL toolchain (Figure~\ref{fig:sil_architecture}).

\subsection{Simulation Results}

\begin{figure}[ht] 
    \centering
    \includegraphics[width=0.82\columnwidth]{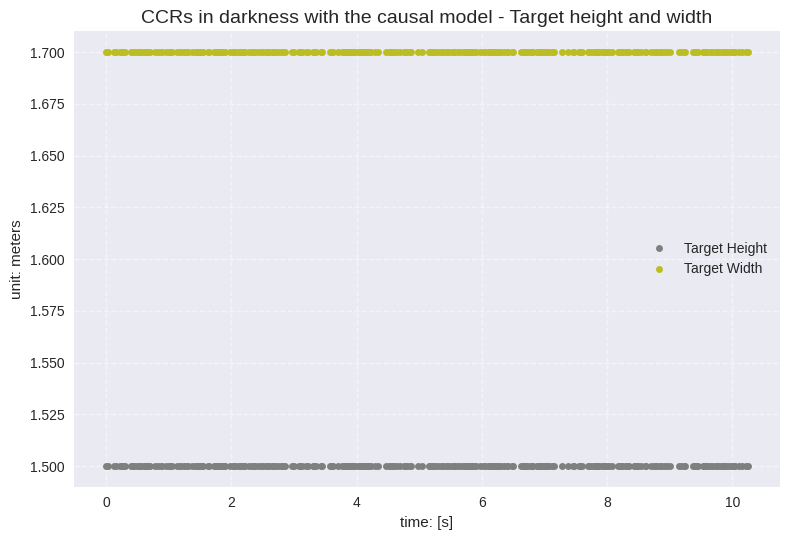}
    \caption{CCRs with the model - target height and width.}
    \label{fig:ccrs_target_height_width}

    \vspace{1em} 

    \includegraphics[width=0.82\columnwidth]{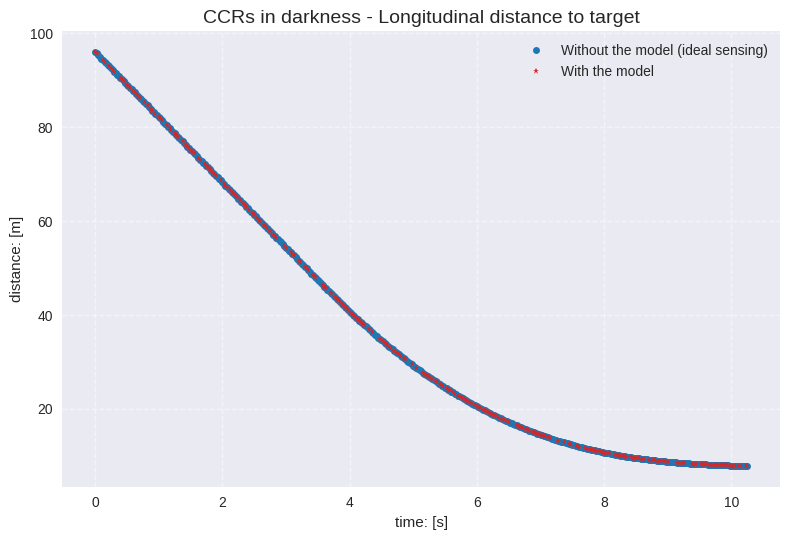}
    \caption{CCRs longitudinal distance to target comparison.}
    \label{fig:ccrs_ego_target_distance}

    \vspace{1em} 

    \includegraphics[width=0.82\columnwidth]{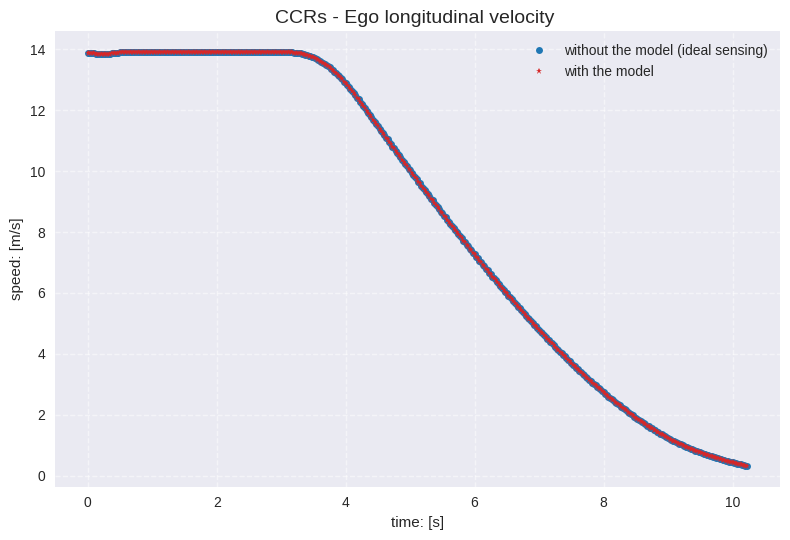}
    \caption{CCRs ego longitudinal velocity comparison.}
    \label{fig:ccrs_ego_speed}

    \includegraphics[width=0.82\columnwidth]{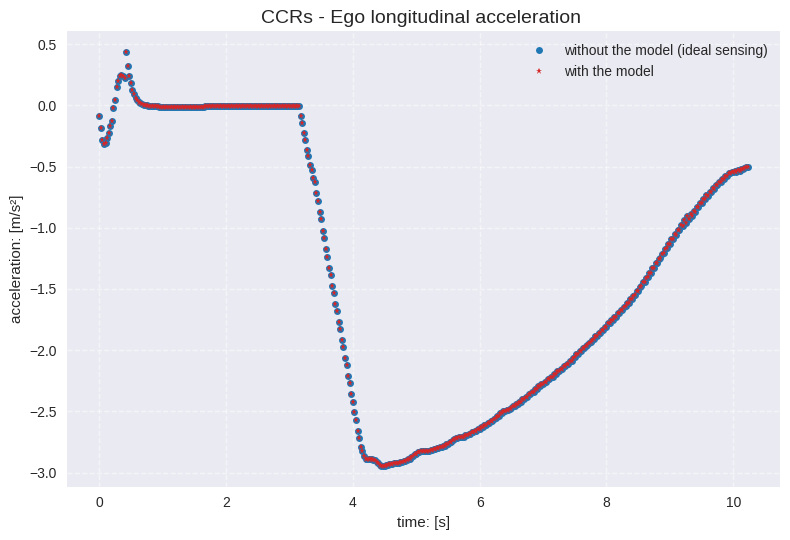}
    \caption{CCRs ego longitudinal acceleration comparison.}
    \label{fig:ccrs_ego_acceleration}
\end{figure}

In the CCRs in Darkness scenario, Fig.~\ref{fig:ccrs_target_height_width}-~\ref{fig:ccrs_ego_acceleration} illustrates the effects of the injected misdetection on the target perception and subsequent ego vehicle behavior. The plot in Fig.~\ref{fig:ccrs_target_height_width} for target height and width reveals distinct gaps representing stochastic misdetections occurring when the darkness-induced noise reduces the Contrast-to-Noise Ratio (CNR) below the required sufficiency threshold. The longitudinal distance to the target (Fig.~\ref{fig:ccrs_ego_target_distance}) shows the comparison between ideal sensing and the causal model. While the ideal sensor maintains a continuous tracking line, the model with injected failures reflects the intermittent loss of the target object due to the aforementioned misdetections. Despite the intermittent perception failures, the longitudinal velocity and acceleration plots (Fig.~\ref{fig:ccrs_ego_speed} and~\ref{fig:ccrs_ego_acceleration}) indicate that the example ADAS remains largely unaffected in this specific instance. The ego vehicle maintains a deceleration profile nearly identical to the ideal sensing case, suggesting that the ADAS control logic possesses sufficient inertia or filtering to handle short-duration perception dropouts.

\begin{figure}[ht]
    \centering
    \includegraphics[width=0.82\columnwidth]{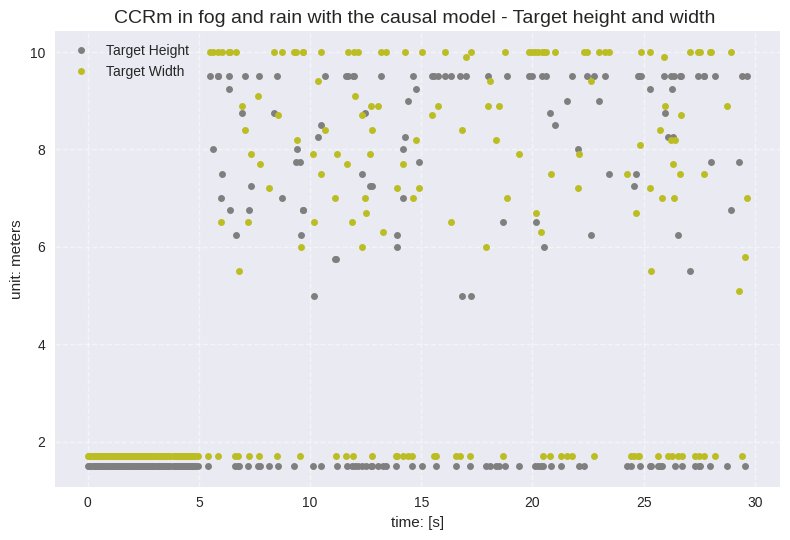}
    \caption{CCRm with the model - target height and width.}
    \label{fig:ccrm_target_height_width}

    \vspace{1em} 

    \includegraphics[width=0.82\columnwidth]{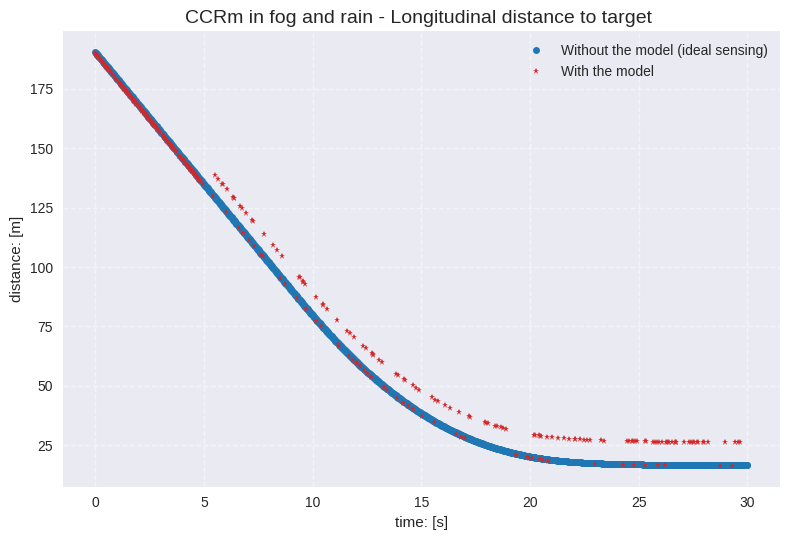}
    \caption{CCRm longitudinal distance to target comparison.}
    \label{fig:ccrm_ego_target_distance}

    \vspace{1em} 

    \includegraphics[width=0.82\columnwidth]{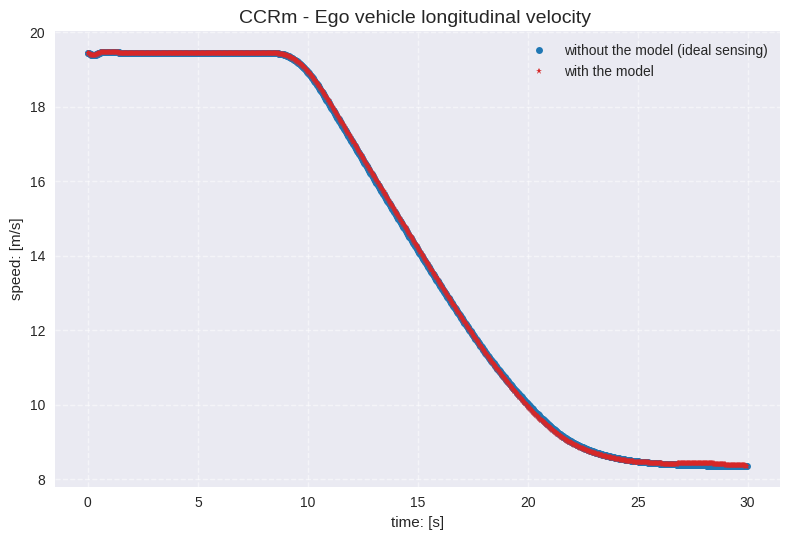}
    \caption{CCRm ego longitudinal velocity comparison.}
    \label{fig:ccrm_ego_speed}

    \vspace{1em} 

    \includegraphics[width=0.82\columnwidth]{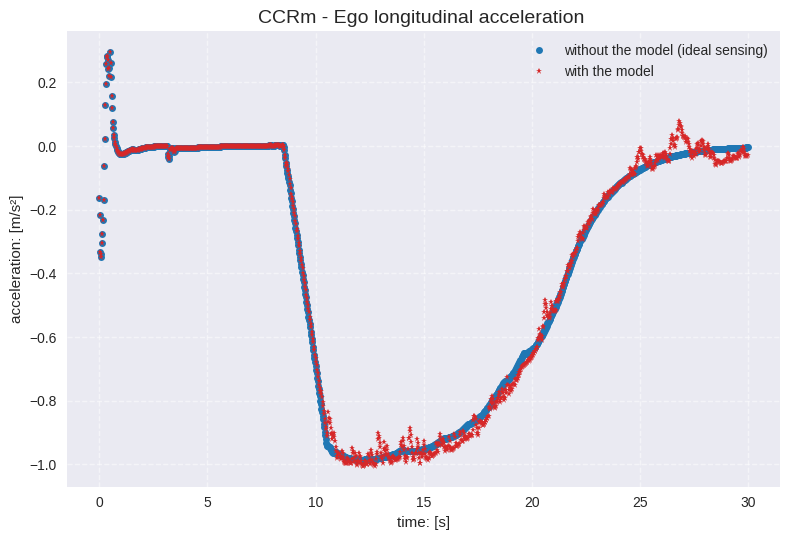}
    \caption{CCRm ego longitudinal acceleration comparison.}
    \label{fig:ccrm_ego_acceleration}
\end{figure}

The CCRm in fog and rain scenario evaluates the cumulative impact of atmospheric scattering and lens-level stressors on perception continuity and localization. The target height and width plot (Fig.~\ref{fig:ccrm_target_height_width}) reveals significant fluctuations and noise in the perceived dimensions of the lead vehicle compared to ground truth. These errors are primarily driven by the sharpness reasoning path, where fog and rain-induced spray degrade object boundaries. The longitudinal distance plot (Fig.~\ref{fig:ccrm_ego_target_distance}) indicates that while the overall trend follows the ideal sensing baseline, the injected model introduces frequent, short-duration misdetections and distance jitter. This is particularly evident as the ego vehicle approaches the target and the relative impact of spray increases. Similar to the CCRs scenario, the ego vehicle's longitudinal velocity (Fig.~\ref{fig:ccrm_ego_speed}) remains almost identical to the ideal sensing case. This suggests that the ADAS speed control logic is robust enough to filter out the high-frequency perception noise injected by the model in this specific following maneuver. In contrast to the stable velocity, the longitudinal acceleration plot (Fig.~\ref{fig:ccrs_ego_acceleration}) shows significant noise and oscillation compared to the smooth profile of the ideal sensor.

\begin{figure}[ht]
    \centering
    \includegraphics[width=0.82\columnwidth]{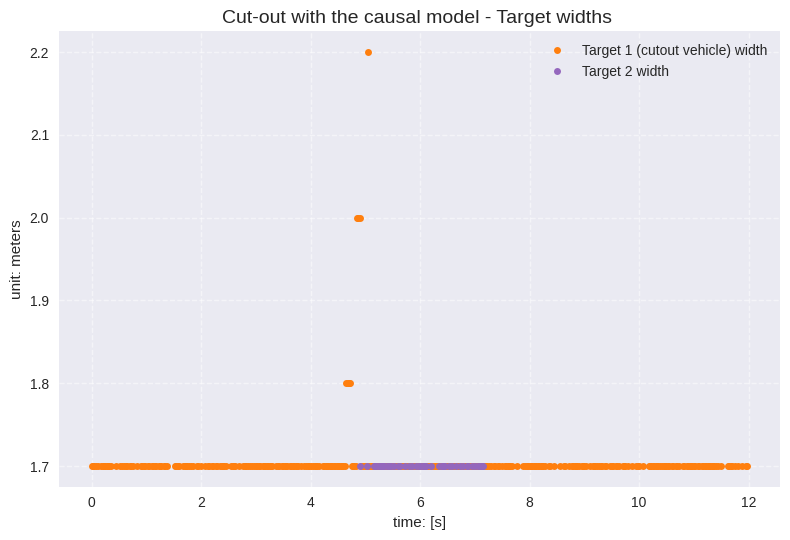}
    \caption{Cut-out with the model - target widths.}
    \label{fig:cutout_target_widths}

    \vspace{1em} 

    \includegraphics[width=0.82\columnwidth]{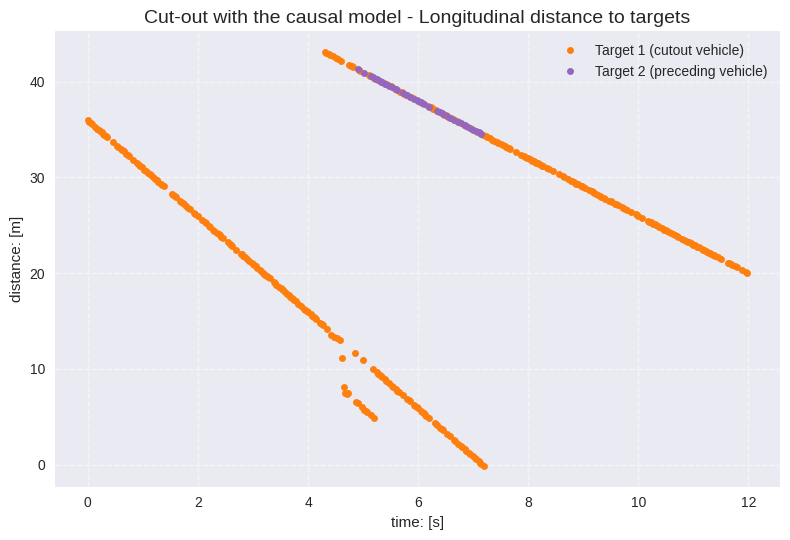}
    \caption{Cut-out longitudinal distance to target comparison.}
    \label{fig:cutout_ego_target_distance}

    \vspace{1em} 

    \includegraphics[width=0.82\columnwidth]{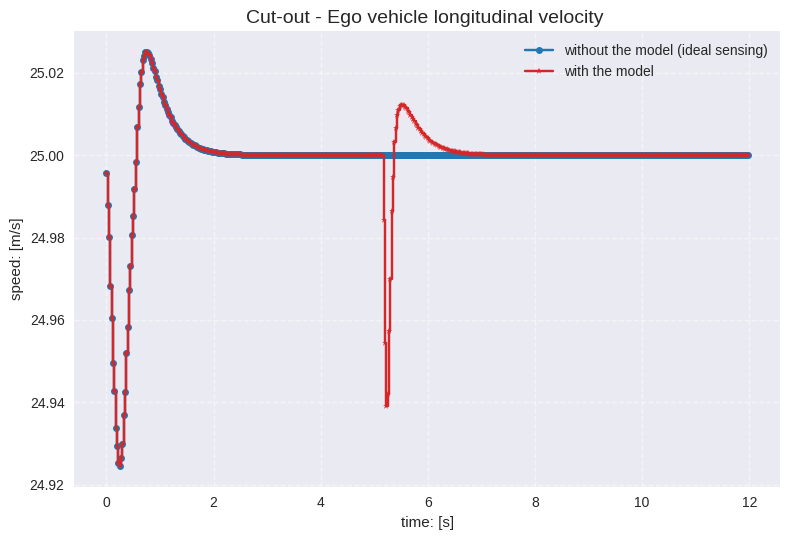}
    \caption{Cut-out ego longitudinal velocity comparison.}
    \label{fig:cutout_ego_speed}

    \vspace{1em} 

    \includegraphics[width=0.82\columnwidth]{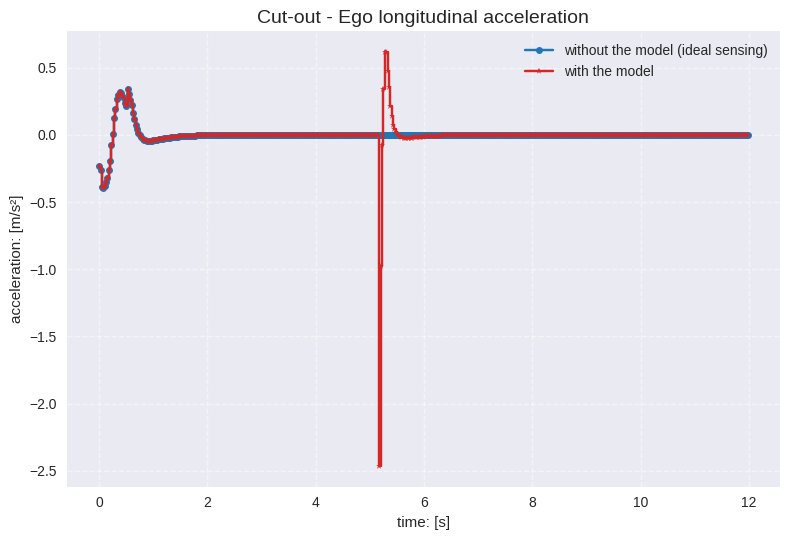}
    \caption{Cut-out ego longitudinal acceleration comparison.}
    \label{fig:cutout_ego_acceleration}
\end{figure}

The Cut-out scenario focuses on evaluating the object merge, as shown in Fig.~\ref{fig:cutout_target_widths} to \ref{fig:cutout_ego_acceleration}. Specifically, the target width plot (Fig.~\ref{fig:cutout_target_widths}) illustrates the activation of the merge logic during the cut-out maneuver. As the overlap and IoU between the two preceding targets increase, the model stochastically replaces the individual vehicle detections with a single, significantly wider merged object. This merging behavior results in several outliers in the longitudinal distance estimation (Fig.~\ref{fig:cutout_ego_target_distance}). When the objects are perceived as a single entity, the perceived distance to the target to appear shorter than the actual ground truth. The impact on the ADAS longitudinal control is evident in the ego vehicle's velocity and acceleration profiles (Fig.~\ref{fig:cutout_ego_speed} and ~\ref{fig:cutout_ego_acceleration}). Unlike the previous scenarios where the function remained stable, the object merge causes a noticeable drop in ego speed and a corresponding deceleration spike. This demonstrates how a perception-level geometric ambiguity can propagate into an unintended and potentially unsafe braking intervention.

\section{Conclusion}
This paper presented a causal model for reasoning about camera-based perception failures, bridging the gap between physical triggering conditions and functional ADAS insufficiencies. By integrating this model into a standardized SIL toolchain, we enabled ADAS robustness evaluation beyond the limitations of ideal-sensor data. This work extends the research on the correlation between physical test tracks and SIL simulations~\cite{fei2024sil}. While the presented correlation study established a baseline using ideal sensing, the causal model allows for realistic perception artifacts that better mirror real-world complexities. Accounting for these stochastic failures significantly improves the correlation between virtual results and physical behavior, providing a more representative environment for safety assurance. Future work will focus on expanding the causal network to include stressors like glare and lens occlusion, as well as calibrating the model against real-world sensor data to ensure failure probabilities align with specific hardware performance.

\section*{Acknowledgment}
The authors would like to thank Andreas Tingberg, Andrew Backhouse and Ludvig Källström for the valuable discussions and Simon Lundell and Emil Knabe for the support on esmini throughout the project.

\bibliographystyle{IEEEtran}
\bibliography{references}

\end{document}